\pdfoutput=1
\documentclass[11pt]{article}

\usepackage[preprint]{acl} 

\usepackage{times}
\usepackage{latexsym}
\usepackage[T1]{fontenc}
\usepackage[utf8]{inputenc}
\usepackage{microtype}
\usepackage{inconsolata}
\usepackage{graphicx}
\usepackage{booktabs}
\usepackage{multirow}
\usepackage{amsmath}
\usepackage{xcolor}
\usepackage{hyperref}

\title{\includegraphics[width=0.04\linewidth]{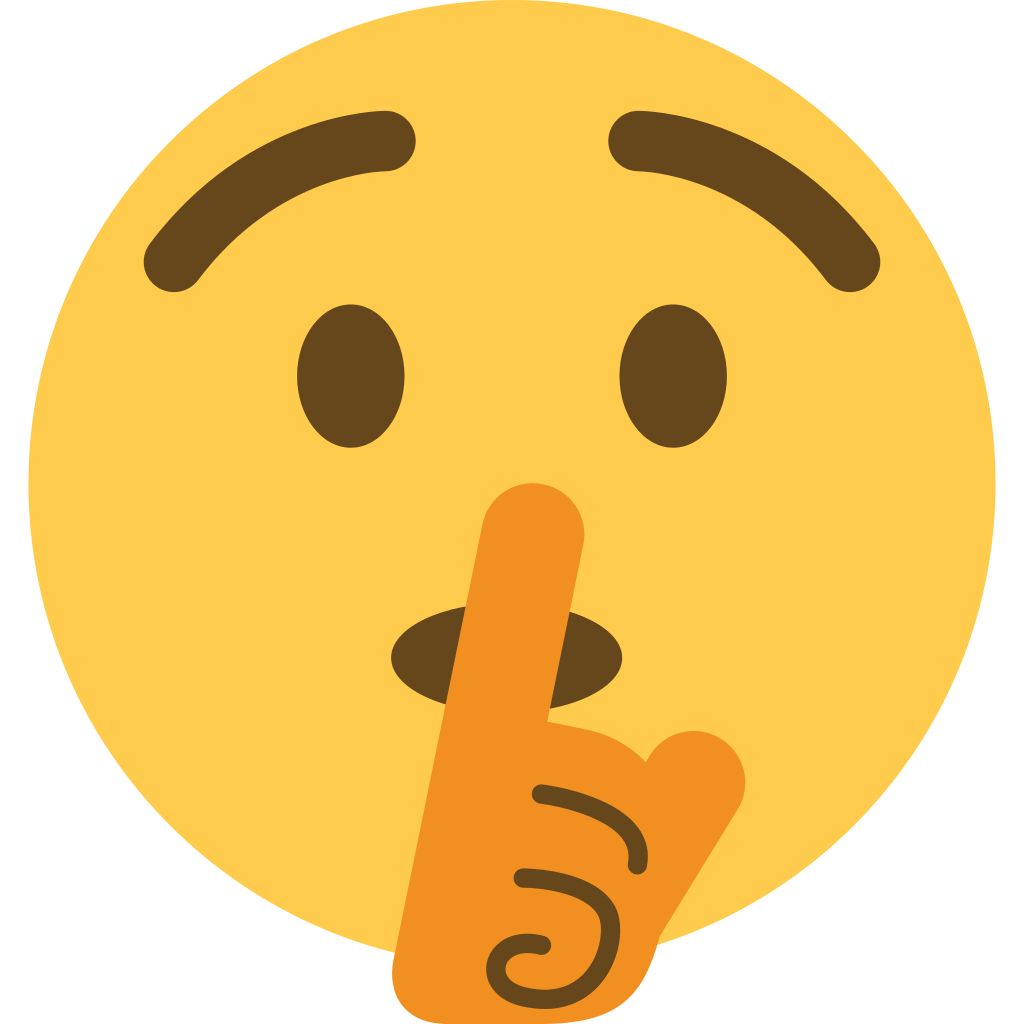} Explain Less, Understand More: Data-Efficient Personalization of Reader-Dependent Jargons}
\author{Bohao Wu$^{\heartsuit}$ \ \quad Qingyun Wang$^{\diamondsuit}$ \ \quad  Yue Guo$^{\heartsuit}$ \\
  $^{\heartsuit}$University of Illinois at Urbana-Champaign \quad $^{\diamondsuit}$ William \& Mary\\
  \texttt{\{bohaowu,yueg\}@illinois.edu} \quad qwang16@wm.edu \\}

\begin{document}
\maketitle

\begin{abstract}
Personalizing jargon detection and explanation is essential for making technical documents accessible to readers with diverse disciplinary backgrounds. However, tailoring models to individual users typically requires substantial annotation efforts and computational resources due to user-specific finetuning.
To address this, we present a systematic study of personalized jargon detection, focusing on methods that are both \emph{efficient} and \emph{scalable} for real-world deployment.
We explore two personalization strategies:
(1) lightweight finetuning using Low-Rank Adaptation (LoRA) on open-source models,
and (2) personalized prompting, which tailors model behavior at inference time without retraining.
To reflect realistic constraints, we also investigate semi-supervised approaches that combine limited annotated data with self-supervised learning from users' publications.
Our personalized LoRA model outperforms GPT-4 with contextual prompting by 21.4\% in F1 score and exceeds the best performing oracle baseline by 8.3\%.
Remarkably, our method achieves comparable performance using only 10\% of the annotated training data, demonstrating its practicality for resource-constrained settings.
Our study systematically explores efficient, low-resource personalization of jargon detection using open-source language models, offering a practical path toward scalable, user-adaptive NLP systems.
\end{abstract}

\section{Introduction}

Large Language Models (LLMs) are increasingly used to support interdisciplinary research by helping scholars navigate diverse and domain-specific texts~\cite{leto-etal-2024-first,ramoneda-etal-2024-role,lu2024llms,jiang2025applications}. However, a persistent barrier to effective interdisciplinary collaboration is the prevalence of domain-specific jargon~\cite{barnett2020growth,Strober2006HabitsOT}. Researchers often struggle to interpret specialized terminology outside their core expertise, leading to miscommunication~\cite{han2018communication,Choi2007MultidisciplinarityIA}, impaired knowledge integration~\cite{lucy2023words}, and ultimately slow scientific discovery~\cite{glasziou2020waste,daniel2022challenges,van2024seven}.

\begin{figure*}[!htbp]
  \includegraphics[trim={0.7cm 0.4cm 0.7cm 0.65cm},clip,width=\textwidth]{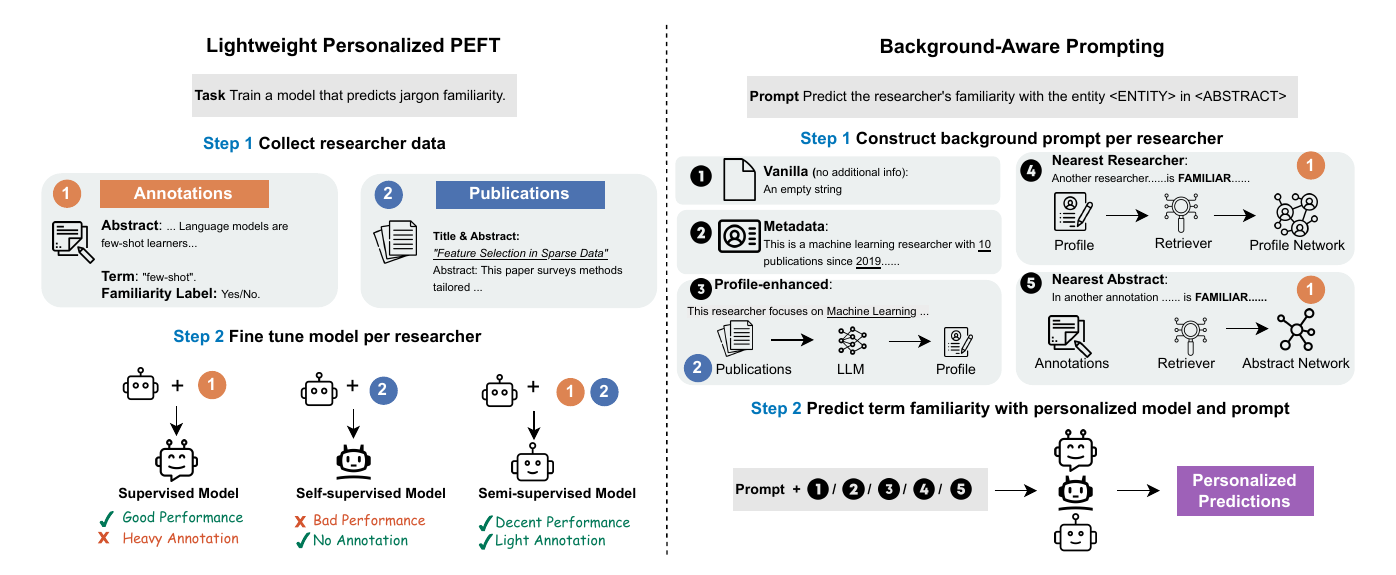}
  \caption {To enable personalized jargon detection, we first fine tune a LoRA-based PEFT model using supervised, self-supervised, and semi-supervised training strategies that reflect real world scenarios with varying levels of annotation availability. Next, we enhance the contextual understanding of the target researcher through a range of background aware prompting methods, including vanilla, metadata based, profile enhanced, nearest researcher, and nearest abstract), to generate personalized familiarity predictions.
  }
  \label{fig:schematic}
\end{figure*}

While prior work has developed NLP methods to identify and simplify scholarly jargon using general-purpose corpora like Wikipedia as proxies for reader knowledge~\citep{10.1093/applin/amt015,TanakaIshii2011WordFA,Guo2022CELLSAP,guo2021automated}, these approaches remain limited by their lack of personalization. A researcher's background significantly influences their familiarity with domain-specific terms~\citep{Gooding2022OneSD,guo-etal-2024-personalized}, suggesting that individualized models could more effectively determine which terms require explanation.

To address this challenge, we focus on the task of personalized jargon identification: automatically detecting domain-specific terms that may be unfamiliar to an individual researcher based on their background. Our goal is to make interdisciplinary content more accessible by leveraging LLMs in a personalized, data-efficient, and scalable manner.

Recent efforts in personalized language models, such as LaMP~\citep{salemi-etal-2024-lamp}, OPPU~\citep{tan-etal-2024-democratizing}, and Per-Pcs~\citep{tan-etal-2024-personalized}, have shown promise by adapting models to user preferences via parameter-efficient fine-tuning (PEFT). However, these methods often rely on costly supervised data or explicit annotation, limiting generalizability. While systems like HLLM~\citep{liang-etal-2024-hllm} target personalization tasks, they do not directly address the broader challenge of personalized language understanding in scholarly contexts.

\citet{guo-etal-2024-personalized} made the first step toward personalized jargon detection by releasing a benchmark and analyzing GPT-4's capabilities. However, their approach depends on costly prompting and rich supervision, raising concerns about scalability and generalization. In this paper, we provide a systematic study of personalized jargon detection with an emphasis on efficiency, scalability, and low-resource practicality. We show that lightweight PEFT on open-source models can surpass GPT-4 while requiring only 10\% annotated data, highlighting the feasibility of scalable, user-adaptive NLP systems.

\section{Method}

To personalize jargon familiarity, we investigate three fine-tuning settings: supervised, self-supervised, and semi-supervised (\S\ref{model_training_main}), and incorporate contextual prompts (\S\ref{prompt_setup_main}). To isolate the impact of personalization, we additionally evaluate a leave-one-annotator-out setting (\S\ref{ablation_main}).

\subsection{Experimental Setup}

We use the personalized jargon detection dataset from \citet{guo-etal-2024-personalized}, which contains 11k term familiarity (i.e., familiar or unfamiliar) annotations provided by 11 computer science researchers for terms extracted from 100 paper abstracts. It is the only interdisciplinary jargon dataset with high-quality annotations, rich personal metadata, and accompanying published papers from the annotators. We follow the same data split as \citet{guo-etal-2024-personalized}, dividing the dataset into 60/20/20 for the train, validation, and test sets. We selected \texttt{Llama-3.1 8B Instruct 4bit}~\citep{dubey2024llama} as our baseline model for personalized jargon detection based on its strong instruction-following ability, low mismatch rate, and competitive performance in preliminary evaluations across several state-of-the-art LLMs. We evaluate model performance using the effective F1 score, which adjusts for output format errors by penalizing predictions with a high mismatch rate. Details of hyperparameters and model selection are in Appendix~\ref{app:setup} and Appendix~\ref{app:model-selection}.

\subsection{Lightweight Personalized PEFT}
\label{model_training_main}

To ensure consistency across different personalization settings, we adopt the Alpaca format~\citep{alpaca}. Each training instance is structured into three components: an \texttt{Instruction}, an \texttt{Input} (target abstract and term), and a \texttt{Response} (binary familiarity label). We use this standardized format to unify model interaction across different training strategies. Task-specific instructions and prompt examples are shown in Appendix Table~\ref{tab:append}. We evaluate three training strategies with varying supervision levels:

\begin{itemize}
  \setlength\itemsep{0pt}
  \setlength\parskip{0pt}
  \setlength\parsep{0pt}
  \item \textit{Supervised}: We adopt LoRA~\citep{hu2021lora} for PEFT, following findings from \citet{tan-etal-2024-democratizing} demonstrating its strong performance. The model is trained to predict annotator familiarity from each term and its associated abstract.
  \item \textit{Self-supervised}: To simulate low-resource scenarios, we fine-tune on unlabeled titles and abstracts from each annotator's prior publications using a causal language modeling (next-token prediction) objective. For annotators with $\leq 5$ publications, we augment the corpus with papers from their self-defined subdomain, reflecting practical cases where personalized models rely on domain-relevant but unlabeled content.
  \item \textit{Semi-supervised}: We examine a hybrid approach that combines limited labeled data with the annotator's publication corpus, evaluating how much supervision is needed to balance annotation effort with personalization quality and to guide future annotation strategies.
\end{itemize}

\subsection{Background-Aware Prompting}
\label{prompt_setup_main}

Building on prior work~\cite{guo-etal-2024-personalized,tan-etal-2024-democratizing}, we design prompting strategies that add varying levels of researcher-specific context to the model input (full prompts in Appendix Table~\ref{tab:prompt}), differing in both type and detail of background information:

\begin{itemize}
  \setlength\itemsep{0pt}
  \setlength\parskip{0pt}
  \setlength\parsep{0pt}
  \item \textit{Metadata}: Structured features including the annotator's self-defined subfield (e.g., NLP, computer vision), publication count, average references, year of first publication, and the domain of the current abstract. They serve as lightweight indicators of expertise and familiarity.
  \item \textit{Profile}: Following prior work on user modeling in personalized recommendation~\cite{tan-etal-2024-democratizing}, we use the baseline model to generate a natural language summary of the annotator's research background based on their metadata.
  \item \textit{Nearest Annotator}: We use BM25~\citep{trotman2014improvements} to retrieve the most similar annotator based on profile text, and use their familiarity labels for the most similar terms as proxy input.
  \item \textit{Nearest Abstract}: We retrieve the most similar abstract using BM25 and use the target annotator's familiarity labels for its terms as context.
\end{itemize}

\subsection{Ablation Study}
\label{ablation_main}

To isolate the effect of personalization, we consider two ablation settings. First, we include a \textit{vanilla prompting} setup with no additional contextual information (e.g., no metadata, profiles, or nearest-neighbor retrieval). While still personalized, since the model is fine-tuned on annotator-specific data (supervised, self-supervised, or semi-supervised), this setting removes auxiliary background features. Second, we evaluate a \textit{non-personalized baseline} using a leave-one-annotator-out scheme, where the model is trained on data from all annotators except the held-out one. For comparability with the supervised personalized model, we subsample the training data to match the same number of examples, keeping all other parameters identical.

\begin{figure*}[htbp]  
    \centering
    \begin{minipage}{0.33\textwidth}
        \centering
        \includegraphics[width=\linewidth]{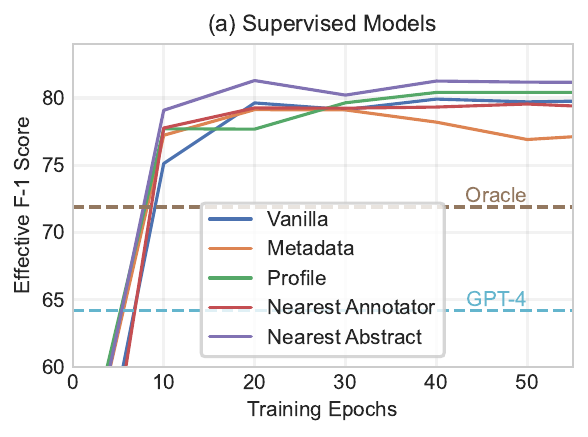}
    \end{minipage}%
    \begin{minipage}{0.33\textwidth}
        \centering
        \includegraphics[width=\linewidth]{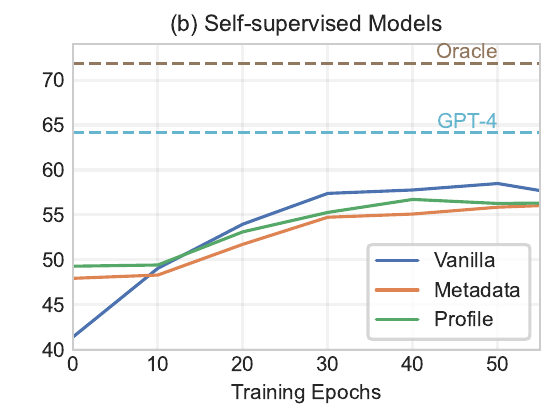}
    \end{minipage}%
    \begin{minipage}{0.33\textwidth}
        \centering
        \includegraphics[width=\linewidth]{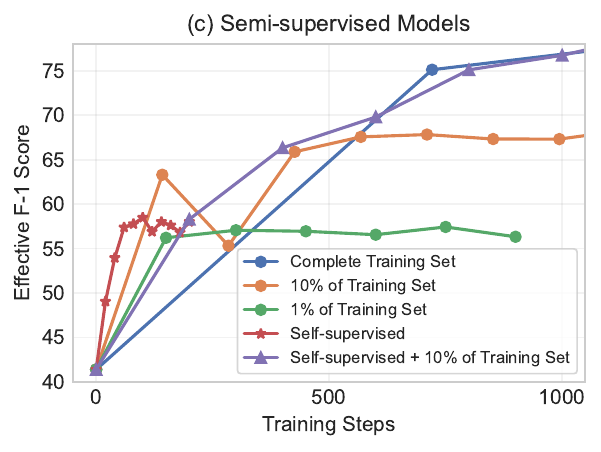}
    \end{minipage}%
    \caption{Validation performance of supervised (a), self-supervised (b), and semi-supervised (c) models for jargon familiarity detection. Each point is a personalized model fine-tuned for one of 11 annotators with different prompting strategies. GPT-4 and Oracle baselines are best results from~\cite{guo-etal-2024-personalized}, where the Oracle uses ratings from the most similar annotator and GPT-4 prompts include five publications. For semi-supervised models, training steps on the x-axis allow consistent comparison across dataset sizes and reflect relative computational cost.}
    \label{fig:super_vs_unsuper}
\end{figure*}

\begin{table}[ht!]
  \centering
  \scriptsize
  \setlength{\tabcolsep}{12pt}
  \renewcommand{\arraystretch}{0.92}
  \begin{tabular}{@{}lccc@{}}
    \toprule
    \textbf{Models} & F1 $\uparrow$ & Recall & Precision \\ \midrule
    \multicolumn{4}{@{}l}{\textbf{Best results from~\cite{guo-etal-2024-personalized}}} \\
    Oracle & $71.9_{\pm 1.7}$ & $76.0_{\pm 2.1}$ & $68.2_{\pm 2.1}$ \\
    GPT-4 & $64.2_{\pm 1.5}$ & $98.7_{\pm 0.5}$ & $47.6_{\pm 1.6}$ \\
    \midrule
    \multicolumn{4}{@{}l}{\textbf{Supervised}} \\
    Vanilla & $\mathbf{77.9}_{\pm 1.2}$ & $77.8_{\pm 2.2}$ & $78.0_{\pm 0.5}$ \\
    Metadata & $76.8_{\pm 1.1}$ & $76.1_{\pm 2.9}$ & $77.7_{\pm 1.0}$ \\
    Profile & $76.6_{\pm 1.0}$ & $73.7_{\pm 2.0}$ & $79.9_{\pm 1.6}$ \\
    Nearest Annotator & $72.1_{\pm 3.0}$ & $70.3_{\pm 4.1}$ & $74.5_{\pm 2.7}$ \\
    Nearest Abstract & $77.8_{\pm 1.1}$ & $78.3_{\pm 1.7}$ & $77.5_{\pm 2.5}$ \\
    \midrule
    \multicolumn{4}{@{}l}{\textbf{Self-supervised}} \\
    Published Papers & $54.6_{\pm 5.1}$ & $77.5_{\pm 7.6}$ & $45.0_{\pm 0.8}$ \\
    \midrule
    \multicolumn{4}{@{}l}{\textbf{Semi-supervised}} \\
    1\% Sup & $53.5_{\pm 2.8}$ & $56.0_{\pm 4.5}$ & $51.5_{\pm 3.2}$ \\
    10\% Sup & $63.6_{\pm 2.9}$ & $59.8_{\pm 5.1}$ & $69.0_{\pm 1.7}$ \\
    Self + 10\% Sup & $\underline{77.0}_{\pm 1.1}$ & $78.9_{\pm 2.6}$ & $75.4_{\pm 0.7}$ \\
    \midrule
    \multicolumn{4}{@{}l}{\textbf{Leave-one-annotator-out (no personalization)}} \\
    Sup baseline & $64.7_{\pm 0.8}$ & $63.9_{\pm 1.1}$ & $65.5_{\pm 0.6}$ \\
    \quad + Metadata & $64.3_{\pm 0.9}$ & $62.7_{\pm 1.5}$ & $66.1_{\pm 0.3}$ \\
    \quad + Profile & $64.9_{\pm 0.8}$ & $64.8_{\pm 0.8}$ & $65.0_{\pm 0.7}$ \\
    Self + 10\% Sup & $53.4_{\pm 0.1}$ & $47.8_{\pm 0.2}$ & $60.6_{\pm 0.1}$ \\
    \quad + Metadata & $58.6_{\pm 0.2}$ & $55.2_{\pm 0.3}$ & $62.5_{\pm 0.2}$ \\
    \quad + Profile & $60.4_{\pm 0.4}$ & $64.2_{\pm 0.3}$ & $57.0_{\pm 0.4}$ \\
    \bottomrule
  \end{tabular}
  \caption{Test-set performance. Unless specified, results use vanilla prompting. Oracle and GPT-4 results follow \citet{guo-etal-2024-personalized}; $\pm$ denotes standard deviation across three runs.}
  \vspace{-5pt}
  \label{tab:results}
\end{table}

\section{Results}

\paragraph{Supervised fine-tuning outperforms GPT-4 and oracle settings.}
On the test set (Table~\ref{tab:results}), vanilla prompting with a supervised personalized model yields the highest F1 (77.9), outperforming GPT-4 with contextual prompting by 21.4\% and the oracle baseline by 8.3\%.
Additional prompting strategies (metadata, profile, nearest annotator, nearest abstract) do not improve over vanilla prompting, suggesting that LoRA fine-tuning itself captures most of the relevant personalized information. The limited gains from these strategies may reflect either noise in background features or insufficient dataset size to reveal their benefits.
Overall, these findings highlight the effectiveness of full supervision with PEFT for modeling the link between annotator background knowledge and jargon familiarity, while showing limited added value from more elaborate prompting.

\paragraph{Self-supervised fine-tuning without annotations shows limited effectiveness.}
In the self-supervised setting, models were fine-tuned solely on each annotator's published papers, without familiarity annotations. Although performance improves slightly over time, overall results remain poor, confirming that publication history alone fails to capture familiarity judgments, consistent with prior findings~\cite{haghani2023makes}. We exclude the nearest annotator and nearest abstract settings from this experiment, as they require access to annotated familiarity labels and are therefore not applicable in the self-supervised scenario.

\paragraph{Semi-supervised personalization enables efficient adaptation with minimal supervision.}
Integrating self-supervised user publication data with only 10\% of labeled training data yields an F1 of 77.0, nearly matching fully supervised performance (77.9) and clearly surpassing models trained on the same limited labeled data alone (63.6).
This demonstrates the value of leveraging unlabeled, domain-relevant data to reduce annotation costs while preserving personalization quality, improving the scalability and accessibility of personalized NLP systems in settings where manual annotation is costly or infeasible.

\paragraph{Personalization is necessary to solve the jargon detection task.}
The personalized model achieves an F1 of 77.9, which is a 20.4\% improvement over supervised leave-one-annotator-out testing (64.7) and a 45.9\% improvement over the self-supervised + 10\% supervised setting (53.4).
While background-aware prompting with metadata or profile does not improve performance over vanilla in the personalized setting, it yields gains without personalized annotation: profile improves F1 by 13.1\% (60.4 vs. 53.4) and metadata by 9.7\% (58.6 vs. 53.4).
These results further underscore the importance of personalization for jargon detection and highlight the effectiveness of PEFT under full supervision, with background-aware prompting offering value only when annotations are limited.

\section{Discussion}

Familiarity judgments are inherently reader-dependent rather than interchangeable annotation noise. The leave-one-annotator-out gap shows that a population model cannot recover an individual researcher's familiarity boundary, even after learning the general task from other researchers. At the same time, publication histories and generated profiles remain imperfect proxies: they help under cold start but cannot reliably replace direct feedback. This motivates personalization systems that request limited calibration data, permit user correction, and avoid treating inferred profiles as authoritative representations of knowledge. Additional transfer, annotator-specific, and domain-specific analyses are provided in Appendix~\ref{app:analysis}.

\section{Conclusion}

In this work, we present a practical and cost-effective framework for personalizing jargon detection in scholarly contexts. By fine-tuning lightweight language models using LoRA, our approach achieves significant performance gains over prior high-resource benchmarks. Our systematic investigation reveals that personalized models, grounded in a researcher's unique background, can match or exceed the performance of proprietary models like GPT-4 using only 10\% of the available annotated data. This underscores the practicality of our method in resource-constrained settings where large-scale human annotation is often infeasible.

\section*{Limitations}
The benchmark includes only 11 computer-science researchers and 100 abstracts, so the magnitude of the personalization gap may not generalize to other populations or disciplines. The dataset contains candidate terms rather than evaluating end-to-end term extraction. Reader-specific models may also become stale as expertise changes. Broader longitudinal evaluation and explicit uncertainty calibration are needed before deployment.

\section*{Ethical Considerations}
Professional histories and familiarity judgments can reveal sensitive information about individuals. Systems should use such data with consent, minimize retention, support deletion and correction, and avoid presenting inferred profiles as authoritative. Personalized explanations may also create filter bubbles or uneven access if deployed without user control.

\section*{GenAI Disclosure}
An AI assistant (ChatGPT) was used solely to improve the grammar of the manuscript. The authors reviewed and edited the output as needed and take full responsibility for the content of this publication.


\bibliography{custom}

\appendix
\section{Full Related Work}

\subsection{Jargon Detection and Simplification} 
Traditional jargon identification typically treats "familiarity" as an objective property of a term, utilizing frequency-based metrics or supervised classifiers trained on general-domain proxies like Wikipedia~\citep{10.1093/applin/amt015, guo2021automated}. While these methods aid general audiences, they fail to account for the specialized expertise of scholars, where the boundary of "jargon" is highly subjective. Recent advancements have moved toward identifying "technicality" via contextual embeddings~\citep{Guo2022CELLSAP}, but these remain static and non-personalized. This lack of flexibility is underscored by a recent study on medical jargon detection, which found that transfer learning across technical datasets yields only modest gains due to divergent labeling objectives, confirming the inherent difficulty of a one-size-fits-all approach to scholarly terminology~\citep{papandreou-etal-2025-medical}. Our work, however, shifts the fundamental focus from "what is jargon" to "what is jargon \textit{for this specific user}," acknowledging that linguistic accessibility in science is user-dependent~\cite{Gooding2022OneSD}.

\subsection{Parameter-Efficient Fine-Tuning (PEFT)}
As Large Language Models (LLMs) scale, full-parameter fine-tuning becomes computationally prohibitive. PEFT methods address this by updating a minimal subset of parameters while keeping pre-trained weights frozen. Early approaches like \textit{adapter tuning}~\citep{houlsby2019parameter} introduced learnable layers within feedforward networks, while \textit{prefix} and \textit{prompt tuning}~\citep{li2021prefix, lester2021power} optimized continuous embeddings to steer model behavior. More recent techniques like \textit{(IA)$^3$}~\citep{liu2022few} scale activations within the attention mechanism. 

In this work, we utilize \textit{Low-Rank Adaptation (LoRA)}~\cite{hu2021lora}, which approximates weight updates through low-rank decomposition. The LoRA framework has been extended to handle even more extreme resource and length constraints; for instance, QLoRA ~\cite{dettmers2024qlora} utilizes 4-bit quantization to enable fine-tuning on consumer-grade hardware, while LongLoRA~\cite{chen2024longlora} extends the effective context length for adapting to long-form documents. We prioritize standard LoRA over these variants to maintain optimal training efficiency and performance stability, as our target model and abstract-level context lengths fit within standard GPU memory constraints without the added computational overhead of quantization or context-extension mechanisms.

\subsection{Personalization and User Modeling}
Personalization has gained traction across several NLP benchmarks, most notably the LaMP benchmark~\citep{salemi-etal-2024-lamp}. Recent frameworks like \textit{One PEFT Per User (OPPU)}~\citep{tan-etal-2024-democratizing} utilize personalized PEFT modules to store user-specific preferences, while \textit{Personalized Pieces (Per-Pcs)}~\citep{tan-etal-2024-personalized} introduces a modular, collaborative approach to enhance scalability. Building on this, \citet{qiu-etal-2025-measuring} recently proposed Difference-aware Personalization Learning (DPL), which emphasizes distilling the unique preference information that distinguishes one individual from another by comparing historical user data.

Despite these advances, existing methods primarily rely on high-resource supervised fine-tuning (SFT), which limits scalability in low-resource scholarly settings. Furthermore, while systems like \textit{HLLM}~\citep{liang-etal-2024-hllm} explore hierarchical personalization, they do not address the specific semantic barriers of technical jargon. We extend this line of research by investigating how semi-supervised and self-supervised PEFT can leverage a researcher’s publication history—an implicit, non-annotated data source—to achieve high-quality personalization without the need for extensive manual labeling.

\section{Full Methodological Details}
\label{app:setup}

To personalize jargon familiarity, we investigate three fine-tuning settings: supervised, self-supervised, and semi-supervised (~\S\ref{model_training}), and incorporate contextual prompts (~\S\ref{prompt_setup}). 
To isolate the impact of personalization, we additionally evaluate a leave-one-annotator-out setting (~\S\ref{ablation}).

\subsection{Experimental Setup}
We utilize the personalized jargon detection dataset from \citet{guo-etal-2024-personalized}, which formalizes the task as an interaction within a \textbf{Personalized Knowledge Space} $\mathcal{K} = \mathcal{J} \times \mathcal{R}$. Here, $\mathcal{J}$ represents the set of technical jargons and $\mathcal{R}$ denotes the set of individual researchers. The dataset contains 11k binary familiarity annotations provided by 11 computer science researchers for terms extracted from 100 paper abstracts. 

Each interaction $(j, r) \in \mathcal{K}$ is augmented by a context function $\mathcal{C}(\cdot)$, which projects the entities into a high-dimensional attribute space. We define the augmented representation of an interaction as:
\begin{equation}
    \mathbf{x}_{j,r} = \left[ \mathcal{C}(j) \oplus \mathcal{C}(r) \right]
\end{equation}
where $\mathcal{C}(j)$ encapsulates jargon-specific metadata (e.g., domain, source abstract) and $\mathcal{C}(r)$ represents the researcher's professional background (e.g., publication history, citation counts). The operator $\oplus$ denotes a concatenation of these contextual features. 

In this work, our objective is to learn a mapping $f: \mathcal{K} \to \mathcal{Y}$ where $\mathcal{Y} = \{``0", ``1"\}$ is the set of target string tokens. Specifically, the model must output the label $y$ such that:
\begin{equation}
\begin{aligned}
y &= f(\mathbf{x}_{j,r}; \Theta), \\
y &\in 
\begin{cases} 
``0" & \text{if } r \text{ is familiar with } j \\
``1" & \text{if } r \text{ is unfamiliar with } j
\end{cases}
\end{aligned}
\end{equation}

where $\Theta$ represents the parameters of the personalized models. This formulation treats personalization as a direct text-to-text generation task conditioned on the joint knowledge space.

We follow the data split established by \citet{guo-etal-2024-personalized}, dividing the interaction pairs into 60/20/20 for the train, validation, and test sets. For our base model, we selected \texttt{Llama-3.1 8B Instruct 4bit}~\cite{dubey2024llama}. This choice was informed by preliminary evaluations across several state-of-the-art LLMs, where \texttt{Llama-3.1 8B Instruct} demonstrated the strongest instruction-following ability and the most competitive baseline performance, which is discussed in Appendix~\ref{app:model-selection}.

\paragraph{Evaluation Metrics}
Our primary objective is to predict the discrete familiarity label $y$ for a given jargon-researcher pair. While we report the standard F-1 score, we also address the practical challenge of model reliability. During our initial baseline selection, we observed that certain models failed to strictly adhere to the binary output format. To account for this, we utilize the \textbf{Effective F-1 Score}, which scales the classification performance by the model's reliability:
\begin{equation}
    \text{eff. F-1} = (1 - \text{Mis. Rate}) \times \text{F-1}
\end{equation}
where the ``Mismatch Rate'' represents the proportion of model outputs that did not conform to the expected format.


\subsection{Lightweight Personalized PEFT}\label{model_training}
To ensure consistency across different personalization settings, we adopt the Alpaca format~\citep{alpaca}, which is widely used in instruction fine-tuning. Each training instance is structured into three components: an \texttt{Instruction}, an \texttt{Input} (target abstract and term), and a \texttt{Response} (binary familiarity label). We use this standardized format to unify model interaction across different training strategies. Task-specific instructions and prompt examples are shown in Table~\ref{tab:append}.
We evaluate three training strategies with varying supervision levels:
\begin{itemize}
\item \textit{Supervised}: We adopt LoRA~\citep{hu2021lora} for PEFT, following findings from \citet{tan-etal-2024-democratizing} demonstrating its strong performance. The model is trained to predict annotator familiarity from each term and its associated abstract.
\item \textit{Self-supervised}: To simulate low-resource scenarios, we fine-tune on unlabeled titles and abstracts from each annotator’s prior publications using a causal language modeling (next-token prediction) objective. For with $\leq 5$ publications, we augment the corpus with papers from their self-defined subdomain, reflecting practical cases where personalized models rely on domain-relevant but unlabeled content.
\item \textit{Semi-supervised}: We examine a hybrid approach that combines limited labeled data with the annotator’s publication corpus, evaluating how much supervision is needed to balance annotation effort with personalization quality and to guide future annotation strategies.
\end{itemize}

\paragraph{Implementation Details}
We fine-tuned the unsloth/Meta-Llama-3.1-8B-Instruct-bnb-4bit model with a maximum sequence length of 2048 tokens. For parameter-efficient training, we applied LoRA with rank 16, scaling factor ($\alpha$) 16, dropout 0, and targeted the projection modules (\texttt{q\_proj}, \texttt{k\_proj}, \texttt{v\_proj}, \texttt{o\_proj}, \texttt{gate\_proj}, \texttt{up\_proj}, and \texttt{down\_proj}). The model was trained for 100 epochs with a per-device batch size of 2 and gradient accumulation of 4 steps, using a learning rate of 2e-4, weight decay of 0.01, and the AdamW (8-bit) optimizer with a linear scheduler and 5 warmup steps. Training was conducted in FP16 or BF16 precision depending on hardware support. Checkpoints were saved every \texttt{epoch\_size * 10 // 8} steps (approximately every 10 epochs).

\subsection{Background-Aware Prompting}\label{prompt_setup}

Building on prior work~\cite{guo-etal-2024-personalized, tan-etal-2024-democratizing}, we design prompting strategies that add varying levels of researcher-specific context to the model input (full prompts in Table~\ref{tab:prompt}), differing in both type and detail of background information:
\begin{itemize}

\item \textit{Metadata}: Structured features including the annotator’s self-defined subfield (e.g., NLP, computer vision), publication count, average references, year of first publication, and the domain of the current abstract. They serve as lightweight indicators of expertise and familiarity.
\item \textit{Profile}: Following prior work on user modeling in personalized recommendation~\cite{tan-etal-2024-democratizing}, we use the baseline model to generate a natural language summary of the annotator’s research background based on their metadata.
\item \textit{Nearest Annotator}: We use BM25~\cite{trotman2014improvements} to retrieve the most similar annotator based on profile text, and use their familiarity labels for the most similar terms as proxy input.
\item \textit{Nearest Abstract}: We retrieve the most similar abstract using BM25 and use the target annotator’s familiarity labels for its terms as context.
\end{itemize}

\subsection{Ablation Study} \label{ablation}
To isolate the effect of personalization, we consider two ablation settings. First, we include a \textit{vanilla prompting} setup with no additional contextual information (e.g., no metadata, profiles, or nearest-neighbor retrieval). While still personalized, since the model is fine-tuned on annotator specific data (supervised, self-supervised, or semi-supervised), this setting removes auxiliary background features. Second, we evaluate a \textit{non-personalized baseline} using a leave-one-annotator-out scheme, where the model is trained on data from all annotators except the held-out one. For comparability with the supervised personalized model, we subsample the training data to match the same number of examples, keeping all other parameters identical.

\section{Baseline Model Selection}
\label{app:model-selection}

\begin{figure}[!htbp]
    \centering
    \includegraphics[width=1\linewidth]{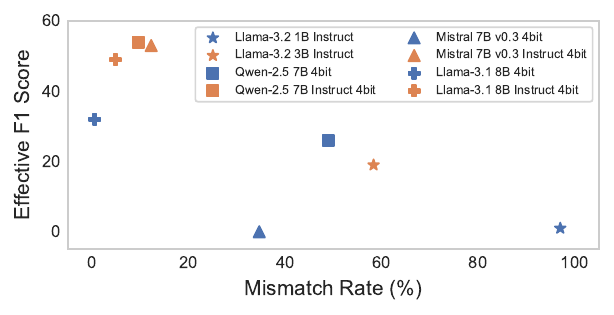}
    \caption{Evaluation results for model selection. Here the inference is done with listed entities to ensure whether the model understands the question.The evaluations are based on the entire dataset. }
    \label{fig:model_selection}
\end{figure}

To establish a robust foundation for our personalized jargon identification task using Parameter Efficient Fine-Tuning (PEFT), we first selected a suitable open-source Large Language Model (LLM) as our baseline. We evaluated several state-of-the-art options, including Llama-2~\citep{touvron2023llama}, Llama-3~\citep{dubey2024llama}, Mistral v0.3~\citep{jiang2023mistral}, and Qwen-2.5~\citep{qwen2,qwen2.5}, considering both their base and instruction-tuned versions in 4-bit and full precision.

For our implementation, we use the \texttt{unsloth} library with all parameters set to their default values, including \texttt{is\_bfloat16\_supported}. The results of this initial evaluation (depicted in Figure~\ref{fig:model_selection}) revealed significant performance variations in terms of both F-1 score and mismatch rate. While Qwen-2.5 7B Instruct 4bit achieved the highest Effective F-1 Score (0.54), and Llama-3.1 8B 4bit exhibited the lowest mismatch rate (0.5\%), we ultimately selected Llama-3.1 8B Instruct 4bit as our baseline for subsequent fine-tuning experiments. This decision was based on its robust performance (Effective F-1 Score of 0.49) and its demonstrated ability to follow instructions with minimal mismatches, suggesting a strong potential for effective adaptation through PEFT for our personalized jargon identification task.

\begin{figure*}[htbp]  
    \centering
    \begin{minipage}{0.33\textwidth}
        \centering
        \includegraphics[width=\linewidth]{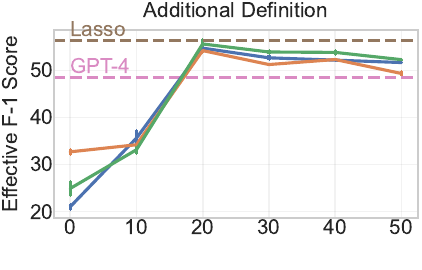}
    \end{minipage}%
    \begin{minipage}{0.33\textwidth}
        \centering
        \includegraphics[width=\linewidth]{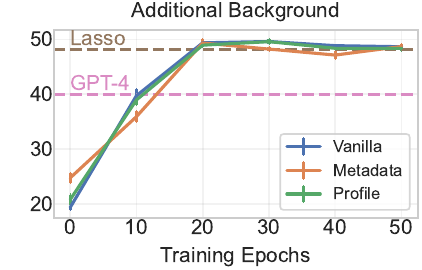}
    \end{minipage}%
    \begin{minipage}{0.33\textwidth}
        \centering
        \includegraphics[width=\linewidth]{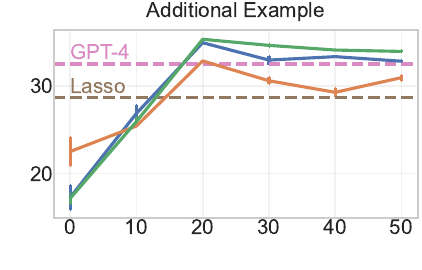}
    \end{minipage}
    \caption{Prediction of additional information needs via various \textbf{Supervised} models fine-tuned on the familiarity annotation data. The results of all three sub-figures are evaluated on the validation set. Here, "Lasso" and "GPT-4" denotes the prediction performance of Lasso regression model and GPT-4, respectively. \citep{guo-etal-2024-personalized}}
    \label{fig:super_add_info_std}
\end{figure*}
\section{Full Discussion and Analysis}
\label{app:analysis}

While our primary results demonstrate the quantitative superiority of personalized PEFT, a robust understanding of user-adaptive systems requires a deeper investigation into their qualitative behavior and generalization capabilities. In this section, we move beyond aggregate metrics to analyze our framework from three critical perspectives. First, we evaluate the robustness of the learned expertise profiles by testing their generalization to unseen downstream information tasks (\S\ref{sec:generalization}). Second, we conduct a granular error analysis of individual researcher profiles to identify the linguistic patterns where personalization offers the most significant gains over baseline models (\S\ref{sec:annotator_analysis}). Finally, we examine the stability of the model's familiarity boundary across disparate technical domains, highlighting the trade-offs between contextual grounding and semantic noise (\S\ref{sec:domain_analysis}).

\subsection{Generalization to Downstream Information Needs} \label{sec:generalization}

To determine if our fine-tuned models have internalized a robust representation of researcher expertise, we evaluate their performance on a structurally distinct generalization task. Beyond binary familiarity, the dataset includes annotations for \textit{Information Requirements}—specific types of auxiliary content a researcher $r$ requires to fully comprehend a jargon $j$. 

We formalize this as a multi-label classification task over the same interaction space $\mathcal{K}$. Let $\mathcal{I} = \{i_{def}, i_{bg}, i_{ex}\}$ represent the set of information needs, corresponding to definitions, background knowledge, and illustrative examples, respectively. Using the previously defined augmented representation $\mathbf{x}_{j,r}$, we evaluate whether the parameters $\Theta$ learned during familiarity training can generalize to the mapping $g$:
\begin{equation}
    \mathbf{y}_{info} = g(\mathbf{x}_{j,r}; \Theta), \quad \mathbf{y}_{info} \in \{0, 1\}^{|\mathcal{I}|}
\end{equation}
where each element in $\mathbf{y}_{info}$ indicates the necessity of a specific information type. 

Figure~\ref{fig:super_add_info_std} illustrates the performance of our models in this zero-shot or low-resource transfer setting. Notably, our fine-tuned LoRA models achieve performance on definition and background knowledge tasks comparable to specialized Lasso regression baselines. Furthermore, our approach significantly outperforms all baselines in predicting the need for illustrative examples ($i_{ex}$). 

These results suggest that the mapping $f$ learned in our primary experiments is not merely a distribution-matching artifact. Instead, the model effectively constructs a \textbf{Latent Expertise Profile} that remains consistent across different modalities of personalized assistance. This demonstrates that by optimizing for $P(y \mid j, r)$, the model captures the underlying semantic boundary between a researcher's current knowledge and the technical requirements of the text.

\begin{figure}[!htbp]
    \centering
    \includegraphics[width=1\linewidth]{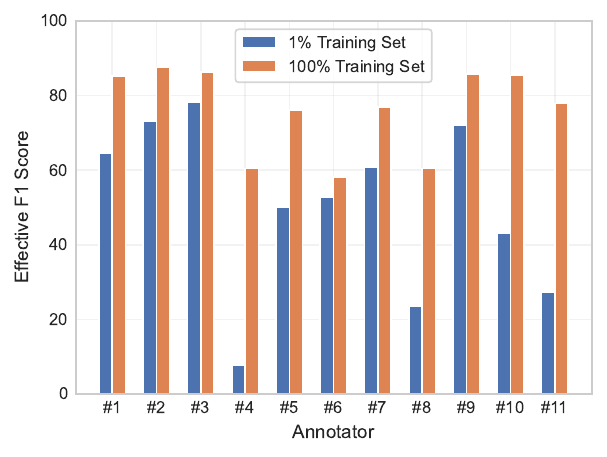}
    \caption{Personalized model Performance for individual annotators.}
    \label{fig:annotator_comparison}
\end{figure}
\subsection{Model Improvement Analysis in Terms of Individual Annotators}\label{sec:annotator_analysis}

To characterize the evolution of the predictive mapping $f(\mathbf{x}_{j,r})$ during fine-tuning, we perform a qualitative error analysis focusing on Annotator \#4—a researcher with a multi-disciplinary background. We compare the baseline model (trained on 1\% of $\mathcal{D}$) against our optimized LoRA-tuned model $\Theta$.

\subsubsection{Baseline Failure Modes} 
The baseline model primarily struggles with regions of the jargon space $\mathcal{J}$ where semantic ambiguity is high. We categorize these failures into four distinct linguistic patterns:
\begin{enumerate}
    \item \textbf{Polysemous Common Words:} Terms that are brief and composed of general vocabulary but carry high specificity in a technical context (e.g., \textit{Radial curves} in Materials Science, \textit{Domains} in Geology, and \textit{Op-amp} in Physics). The model often fails to differentiate these from general language use.
    \item \textbf{Non-Compositional Multi-word Expressions (MWEs):} Complex terms where the meaning is not a straightforward combination of individual words (e.g., \textit{Bayesian optimal mechanism}, \textit{Riemannian framework}, \textit{Bose-Einstein condensate}). The baseline lacks the capacity to capture the deep semantic dependencies within these units.
    \item \textbf{Opaque Acronyms:} Condensed tokens lacking explicit semantic clues (e.g., \textit{CW-SSIM}, \textit{MANOVA}, \textit{ARMAX model}). These require the model to leverage the researcher's background $\mathcal{C}(r)$ more heavily than textual features.
    \item \textbf{Interdisciplinary Overlap:} Terms spanning multiple fields (e.g., \textit{Monolayers}, \textit{Regulatory mechanisms}), which increase the complexity of the familiarity boundary.
\end{enumerate}

\subsubsection{The Recall-Precision Trade-off} 
While the improved model ($\Theta$) demonstrates a higher $F_1$ score and significantly improved recall, we observe a characteristic shift toward \textbf{False Positives} ($y=1$ predictions for familiar terms). This phenomenon occurs primarily with high-frequency methodological or statistical terms such as \textit{Meta-analyses}, \textit{Post-test}, and \textit{Quantitative research}.

This suggests that as the model optimizes for the "unfamiliarity" label to capture a broader range of jargon, it becomes hypersensitive to academic-standard terminology. The model essentially conflates \textit{scholarly frequency} with \textit{personal difficulty}. This indicates a strategic trade-off where the system prioritizes identifying potential knowledge gaps ($y=1$) at the cost of flagging terms that are relatively well-understood within the broader scientific community.

\begin{figure}[!htbp]
    \centering
    \includegraphics[width=1\linewidth]{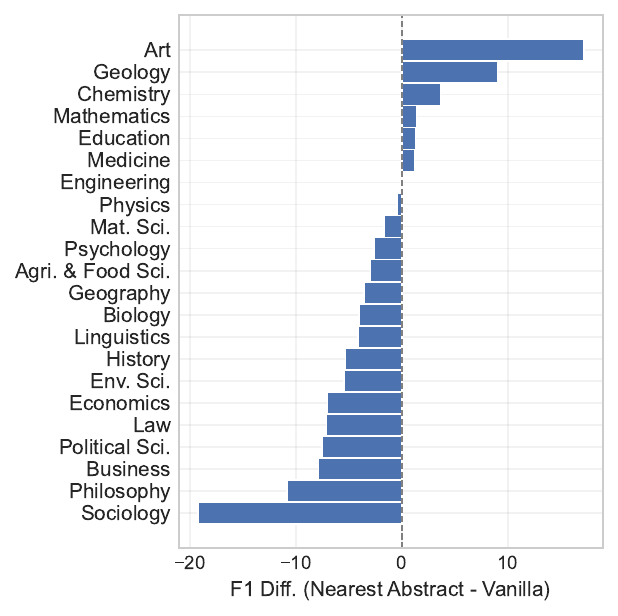}
    \caption{Supervised model performance difference for nearest abstract versus vanilla.}
    \label{fig:category_comparison_sup}
\end{figure}

\subsection{Model Analysis in Terms of Jargon Domain}\label{sec:domain_analysis}

To understand the impact of contextual information on model stability, we compare our two highest-performing configurations: \textbf{Vanilla Prompting} and \textbf{Nearest Abstract (NAb) Prompting}, both trained with 100\% supervision. We analyze their performance across the Art ($\mathcal{J}_{Art} \subset \mathcal{J}$) and Philosophy ($\mathcal{J}_{Phil} \subset \mathcal{J}$) domains to evaluate how the context function $\mathcal{C}(j)$ influences the familiarity boundary.

\subsubsection{The Art Domain: Mitigation of Over-Sensitivity} 
In $\mathcal{J}_{Art}$, the Vanilla model exhibits a significantly higher false positive rate, incorrectly labeling high-frequency terms such as \textit{Reactions} and \textit{Stylistic} as jargon ($y=1$). In contrast, the NAb model—which incorporates the source abstract into $\mathcal{C}(j)$—correctly identifies these as familiar terms. This suggests that without contextual grounding, the model is overly sensitive to terms that possess both a technical use-case and a common-usage baseline. The inclusion of the nearest abstract provides a semantic anchor that prevents the model from misclassifying general vocabulary as domain-specific jargon.

\subsubsection{The Philosophy Domain: Contextual Confusion and Recall Gaps} 
Conversely, in the more abstract domain of $\mathcal{J}_{Phil}$, the NAb model faces a dual challenge in precision and recall:
\begin{enumerate}
    \item \textbf{Precision Decay:} The NAb model demonstrates a tendency toward "Over-Identification," misclassifying foundational philosophical terms like \textit{Structural constraints}, \textit{Analytic philosophers}, and \textit{Argument Facets} as unfamiliar.
    \item \textbf{Recall Deficiency:} Surprisingly, the NAb model fails to detect highly specialized terms such as \textit{Nonmonotonic inference methods}, \textit{Super-knotty rope}, and \textit{Dialectical quality}. 
\end{enumerate}

These results indicate that in highly abstract or linguistic-heavy fields like Philosophy, the context function $\mathcal{C}(j)$ may introduce noise rather than clarity. While the NAb approach excels in grounding concrete terms (as seen in Art), it struggles to model the expertise boundary for complex, nested philosophical concepts. This suggests that for certain subspaces of $\mathcal{J}$, the interaction between jargon and researcher requires a more sophisticated profile of the researcher's specific sub-domain expertise $\mathcal{C}(r)$ rather than just textual proximity to an abstract.

\section{Complete Prompt Configurations}

\begin{table*}
  \centering
  \renewcommand{\arraystretch}{1.5} 
  \small
  \begin{tabular}{p{0.3\linewidth} p{0.65\linewidth}}
    \toprule
    \textbf{Strategies} & \textbf{Related data} \\
    \midrule
    Vanilla & "" (empty string) \\
    Metadata & Self-defined subfield of the reader is: \{\} \\
             & Number of papers published by the reader is: \{\} \\
             & Number of papers referenced by the reader is: \{\} \\
             & Year of the first paper published by the reader is: \{\} \\
             & Domain of study of the paper is: \{\} \\
    Profile-enhancement & This reader is a domain expert in natural language processing (NLP) ... (Machine-generated profile) \\
    Nearest annotator & Another researcher similar to the reader has read the same abstract. For the entity list \{entity\_list\}, this researcher provides the familiarity list as \{familiarity\_list\}. \\
    Nearest abstract & For another similar abstract with the entity list \{entity\_list\}, this reader provides the familiarity list as \{familiarity\_list\}. \\
    \bottomrule
  \end{tabular}
  \caption{The prompting strategies for both supervised fine-tuning and inference.}
  \label{tab:prompt}
\end{table*}

\begin{table*}
  \centering
  \renewcommand{\arraystretch}{1.5} 
  \small
  \begin{tabular}{p{0.12\linewidth} p{0.38\linewidth} p{0.45\linewidth}}
    \toprule
    \textbf{Tasks} & \textbf{Instructions} & \textbf{Prompt} \\
    \midrule
    Familiarity classification & 
    Your job is to estimate how much the reader knows about an entity. You will be provided with the entity, the abstract where the entity came from, and related data about either the reader or the abstract. Your answer should be a list of binary, either 0 or 1, of the same length as the entity list, with no other words. & 
    Entity: \{entity\} \newline
    Abstract: \{abstract\} \newline
    Additional information: \{related\_data\} \newline
    Here's how to gauge the reader's familiarity: \newline
    - 0: The reader knows this subject well and can describe it to others. \newline
    - 1: The reader has either encountered this subject before but knows little about it, or has never come across it at all. \newline
    Based on the information provided, determine familiarity score list corresponding to the entity list, a list of either 0 or 1. \\
    \midrule
    Definition needs classification & 
    You are tasked with predicting whether the reader might need \textbf{additional Definition/Explanation} to fully grasp the entities mentioned in a given abstract. You will be provided with the entity list, the abstract where the entities come from, and related data pertinent to the reader or the abstract. A definition provides key information on the term independent of any context. & 
    Entity: \{entity\} \newline
    Abstract: \{abstract\} \newline
    Additional information: \{related\_data\} \newline
    Provide the estimation whether additional information is needed in a list in the order of the entity. The estimation should be either 0 (no) or 1 (yes). No need to mention the entity. \\
    \midrule
    Background needs classification & 
    You are tasked with predicting whether the reader might need \textbf{additional Background/Motivation} to fully grasp the entities mentioned in a given abstract. Background introduces information important for understanding the term in the context of the abstract. & 
    Entity: \{entity\} \newline
    Abstract: \{abstract\} \newline
    Additional information: \{related\_data\} \newline
    Provide the estimation whether additional information is needed in a list in the order of the entity. The estimation should be either 0 (no) or 1 (yes). No need to mention the entity. \\
    \midrule
    Example needs classification & 
    You are tasked with predicting whether the reader might need \textbf{additional Example} to fully grasp the entities mentioned in a given abstract. An example offers specific instances that help illustrate the practical application or usage of the term within the abstract. & 
    Entity: \{entity\} \newline
    Abstract: \{abstract\} \newline
    Additional information: \{related\_data\} \newline
    Provide the estimation whether additional information is needed in a list in the order of the entity. The estimation should be either 0 (no) or 1 (yes). No need to mention the entity. \\
    \bottomrule
  \end{tabular}
  \caption{The configuration of instructions and prompts for training and inference, following the prompting format from the previous work \citep{guo-etal-2024-personalized}.}
  \label{tab:append}
\end{table*}

\end{document}